# Robust Palm-Vein Recognition Using the MMD Filter: Improving SIFT-Based Feature Matching

Kaveen Perera, Fouad Khelifi, Ammar Belatreche
Computer & Information Sciences department
University of Northumbria at Newcastle, United Kingdom

## Abstract

A major challenge with palm vein images is that slight movements of the fingers and thumb, or variations in hand posture, can stretch the skin in different areas and alter the vein patterns. This can result in an infinite number of variations in palm vein images for a given individual. This paper introduces a novel filtering technique for SIFT-based feature matching, known as the Mean and Median Distance (MMD) Filter. This method evaluates the differences in keypoint coordinates and computes the mean and median in each direction to eliminate incorrect matches. Experiments conducted on the 850nm subset of the CASIA dataset indicate that the proposed MMD filter effectively preserves correct points while reducing false positives detected by other filtering methods. A comparison with existing SIFT-based palm vein recognition systems demonstrates that the proposed MMD filter delivers outstanding performance, achieving lower Equal Error Rate (EER) values.

---

---

## 1. Introduction

The field of biometrics has seen substantial growth recently, driven by increasing global demands for contactless digital security solutions. Biometric characteristics are distinguished by unique features and patterns that enable the identification and verification of individuals, offering a higher level of security compared to traditional methods such as passwords or PIN codes [2]. Palm vein recognition systems operate by analysing the vein structures beneath the skin of the palm. Each individual has a distinct vein pattern that remains relatively stable throughout their lifetime, making vascular biometrics a highly secure and dependable identification method that provides detailed texture information.

One of the key advantages of palm vein recognition technology is that it is contactless, hygienic, non-invasive, and easy to use, all of which contribute to its widespread acceptance [3]. Deoxygenated veins beneath the skin absorb Near-Infrared (NIR) light, causing them to appear darker. However, palm vein images tend to have low contrast and often appear blurry due to the scattering of NIR light by the skin (refer to **Figure 1** [a]). Additionally, image quality is further degraded by sensor noise, which complicates the processes of image processing and feature extraction. Therefore, applying an appropriate contrast enhancement technique before feature extraction is essential. Another challenge with palm vein images is that minor variations in hand posture, such as slight movements of the fingers, thumb, or stretching of the skin, can modify the vein patterns, creating an almost limitless number of variations for the same individual.

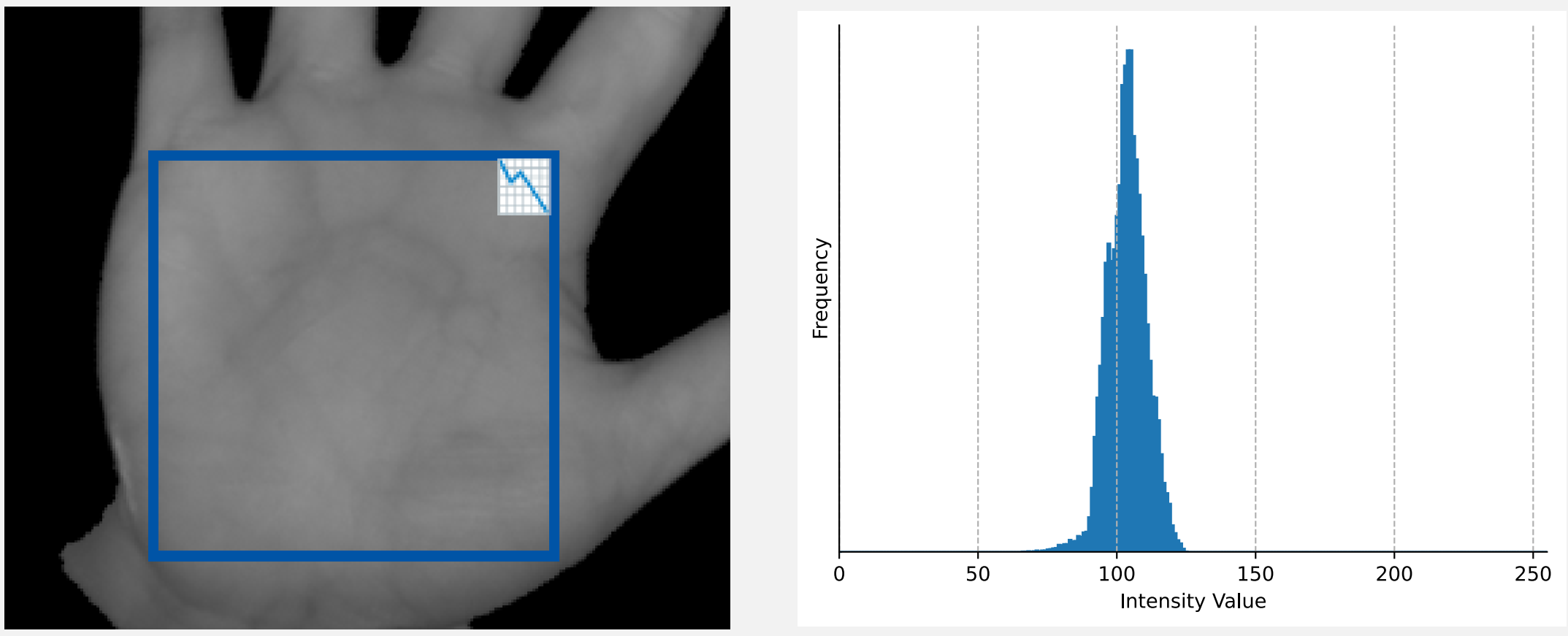

[a]: Sample image from the CASIA dataset [b]: Histogram of [a]

**Figure 1:** Sample image from the CASIA [3] dataset and the corresponding histogram. ROI used to generate the histogram is marked with a blue rectangle.

To address the limitations of current contrast enhancement methods in palm vein imaging, we introduced the Multiple Overlapping Tiles (MOT) method in [4], and later renamed Intensity-Limited Adaptive Contrast Stretching with Layered Gaussian-weighted Overlapping Tiles (ILACS-LGOT) in [5] for clarity and to better reflect the operations involved.

The ILACS-LGOT method has been evaluated within existing palm vein recognition systems that utilise Scale Invariant Feature Transform (SIFT) [4], [5] and RootSIFT [45] features. SIFT is particularly useful for detecting distinctive image features and generating descriptors that can be used to identify matching keypoints between two images. These features are robust against changes in scale, orientation, and position.

This research employs the ILACS-LGOT technique to improve image contrast and investigates SIFT feature matching using different methods, including Euclidean Distance (ED) [6], k-nearest neighbour (KNN) [7], and Lowe's distance ratio test (RT) [8]. Despite these approaches, a significant number of false positives still persist due to the similarity of palm vein features.

To address the issues identified with existing feature matching and filtering methods on palm-vein images, a novel keypoint filtering method, Mean and Median Distance (MMD) Filter is presented. This method considers the mean and median distances between the horizontal and vertical distances of the geometric locations of matched keypoints, then use a set of rules to determine false positives.

The MMD filter when incorporated with the previously evaluated ILACS-LGOT+KNN+RT method, can significantly reduce false matches in SIFT-based feature matching.

The performance of the proposed MMD filter is evaluated using existing SIFT-based palm-vein recognition systems.

The rest of the article is organised as follows: Section 2 thoroughly examines the issues with SIFT feature matching with ED and KNN+RT parameters, while Section 3 introduces the MMD filter. Section 4 presents the evalaution critira followed by results and analysis in Section 5. Section 6 concludes with final remarks and suggestions for future improvements.

The image pair used for the examples is from Subject 001_r of the CASIA [9] dataset. All examples presented in this article have been enhanced using the ILACS-LGOT method.

# 2. SIFT feature matching of palm-vein images

## 2.1 SIFT keypoints

SIFT is a widely used method in computer vision for detecting and describing local features in images. The algorithm is based on the Difference of Gaussians (DoG), which approximates the Laplacian of Gaussian (LoG), a technique to identify regions of interest by detecting local extrema (minima or maxima) in scale-space. This approach is computationally efficient and provides precise localisation of keypoints [8], [10].

The SIFT process begins by successively blurring the input image to construct a Gaussian-scale space. The resulting images, organised into "octaves," are downsampled and blurred again. The adjacent blurred images are subtracted within each octave to generate DoG images, where keypoints are identified as local extrema.

Once potential keypoints are detected, they undergo a localisation process to refine their positions and orientations, ensuring stability against noise and low contrast. Unstable keypoints, particularly those along edges, are discarded by analysing the ratio of principal curvatures from the Hessian matrix.

## 2.2 SIFT Feature Descriptor

To establish the orientation of a keypoint, the magnitudes and directions of the image gradient are first calculated within its surrounding area. The scale of the keypoint is then used to determine the appropriate Gaussian blur to apply to the image. To construct the feature descriptor, the gradient magnitudes and orientations are computed from a 16×16 pixel region surrounding the keypoint.

To ensure the descriptor remains invariant to rotation, a Gaussian weighting function is applied to assign different levels of importance to pixels within this region. The coordinates of the patch are then rotated according to the detected keypoint orientation. The gradient information from 4×4 subregions within this patch is used to generate 16 orientation histograms, each containing 8 orientation bins.

To mitigate the effects of abrupt changes in the image, the gradient information is interpolated into adjacent histogram bins. The final feature descriptor is formed by combining these histograms into a 128-element vector, which is then normalised to unit length to reduce the impact of variations in illumination [11].

## 2.3 Descriptor Matching

For matching, these descriptors are compared using Euclidean Distance (ED), which quantifies similarity by calculating the distance between feature vectors [10]. A keypoint descriptor from image A is compared against all keypoints in image B to find the most similar match. The best match is determined by selecting the pair of keypoints with the smallest ED between their feature descriptor vectors. This process is known as closest-neighbour matching.

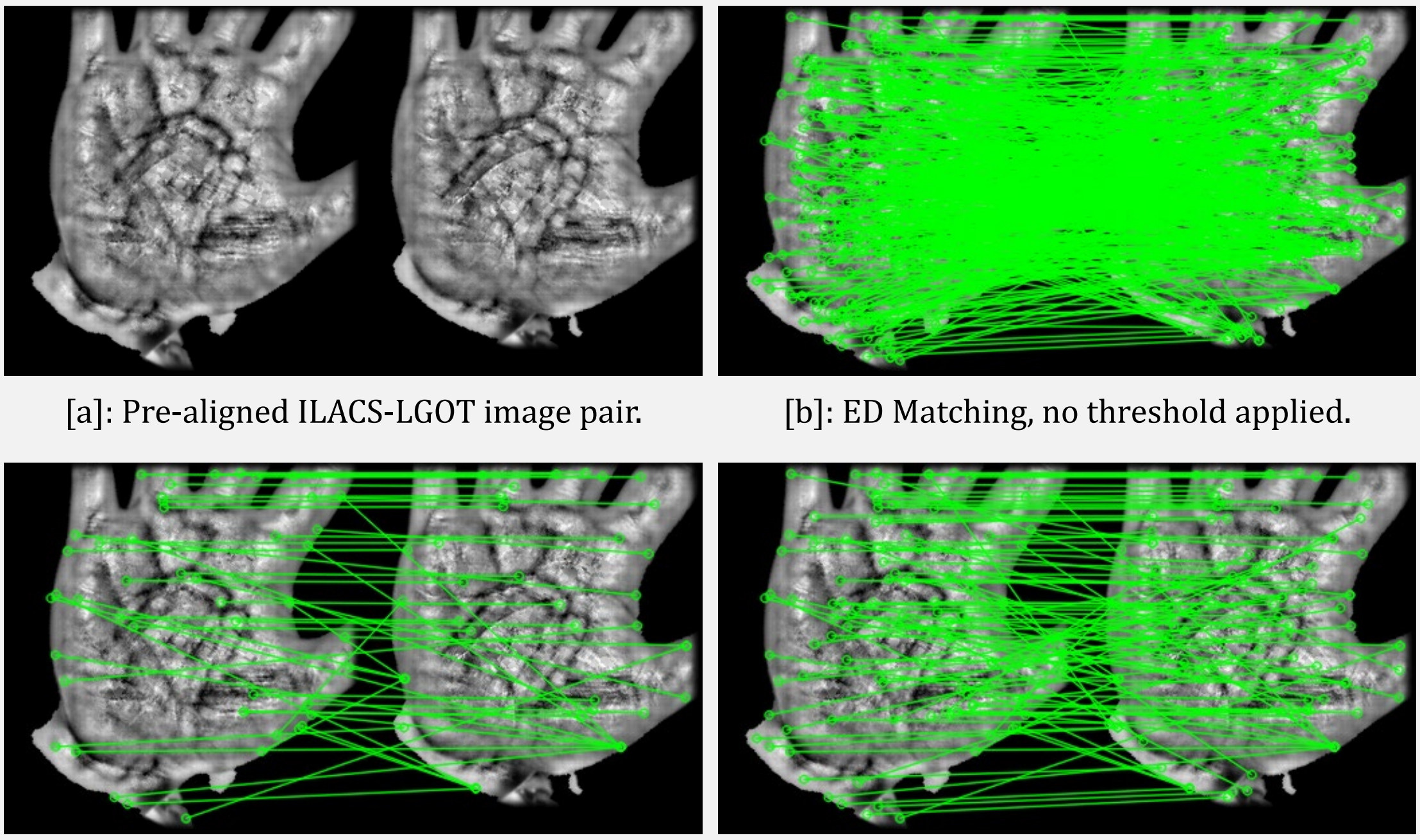

[a]: Pre-aligned ILACS-LGOT image pair. [b]: ED Matching, no threshold applied.

[c]: ED Matching, threshold applied - 250. [d]: ED Matching, threshold applied - 300.

**Figure 2:** Closest-neighbour matching of SIFT features uses the Euclidean Distance (ED) as the distance measure. The keypoints near the edges were retained for demonstration purposes only.

**Figure 2** demonstrates closest-neighbour SIFT matching of two palm images from the same subject, utilising ED. A threshold can be applied for closest-neighbour matching to filter out matches that exceed a set distance, as illustrated in subfigures [c] and [d] of **Figure 2**. However, this method is unreliable, as it may retain numerous incorrect matches.

To refine the matching process and reduce false positives, the distance ratio test (RT) introduced by Lowe [8] adds a layer of precision to the matching strategy. Instead of simply identifying the closest match by the smallest ED, the RT involves calculating the ratio of the distance to the nearest neighbour and the distance to the second-nearest neighbour.

A keypoint match is considered valid if the ratio of the smallest to the second-smallest ED is below a certain threshold, typically set between 0.7–0.8 as suggested by Lowe. This 'Ratio Test' ensures that the selected match is significantly closer than the next best match, thereby increasing the confidence in the robustness of the match By enforcing this condition, the method significantly reduces the likelihood of false matches, ensuring that the matches are not only close but also uniquely closer than any other potential matches.

The most commonly used second closest-neighbour matching technique with SIFT feature matching is the k-nearest neighbours (KNN) algorithm [7]. **Figure 3** demonstrates this matching process using SIFT+KNN+RT with various distance ratios.

RT can filter out almost all the false matches when using lower ratios. In the examples presented in subfigures [a] and [b] of **Figure 3**, a small number of the highest quality matches were retained when using RT values of 0.5 and 0.6, respectively.

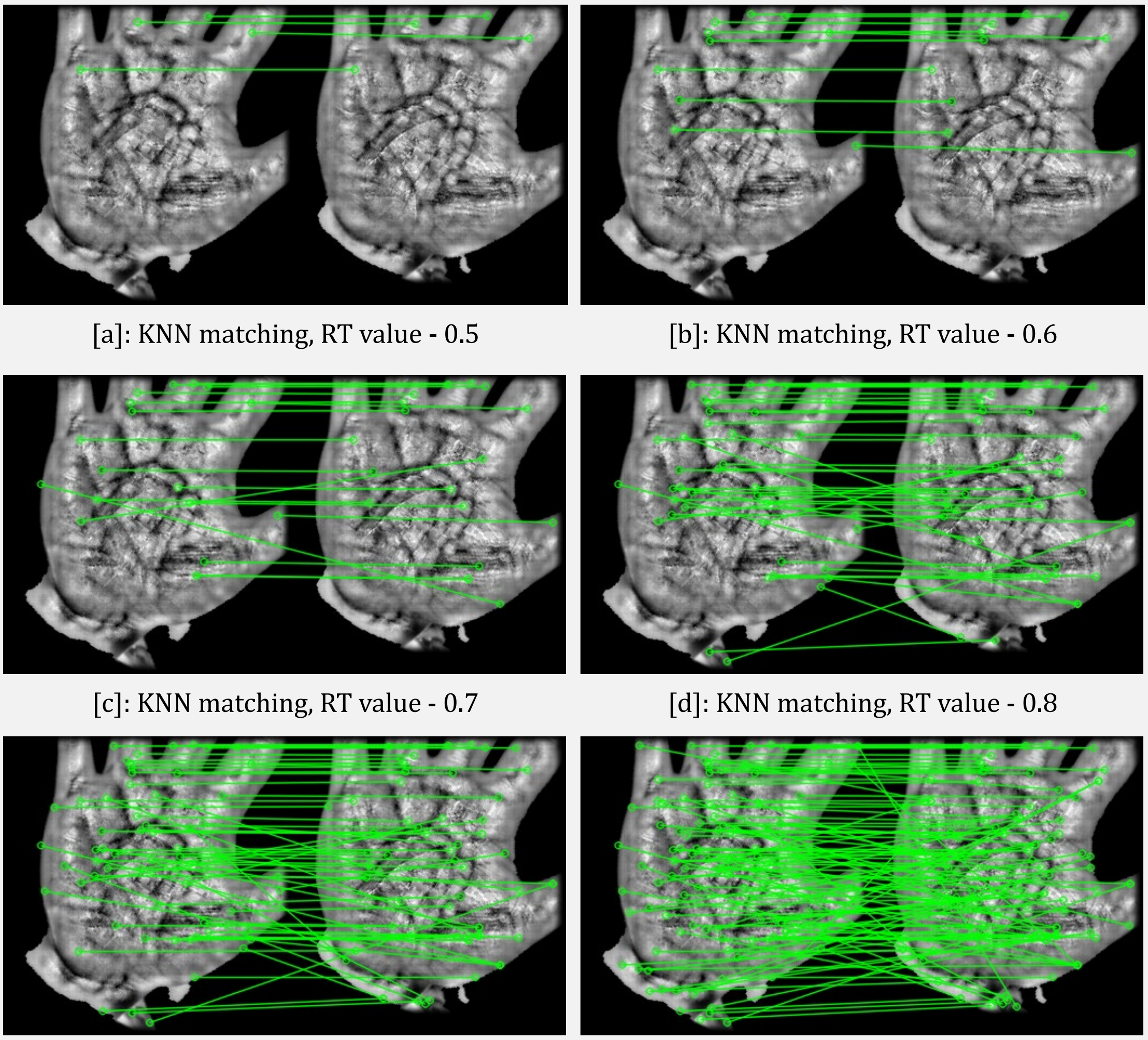

[a]: KNN matching, RT value - 0.5 [b]: KNN matching, RT value - 0.6

[c]: KNN matching, RT value - 0.7 [d]: KNN matching, RT value - 0.8

[e]: KNN matching, RT value - 0.85 [f]: KNN matching, RT value - 0.9

**Figure 3:** k-nearest neighbour (KNN) matching with the Ratio Test (RT) applied. Lower RT values can filter out many useful match pairs, while higher RT values may still preserve a considerable number of false match pairs.

Increasing the RT value, as presented in subfigures [c] through to [f] of **Figure 3**, can start to filter in false matches, and the results gradually begin to appear similar to those of SIFT+ED. However, even with higher RT values, SIFT+KNN+RT can still preserve higher quality matches than SIFT+ED demonstrated in **Figure 2**. However, given the similarity of palm features, these filtering methods still result in a considerable number of false positive matches.

This issue arises because palm-vein images typically offer a narrower feature space, leading to smaller differences between SIFT feature descriptors. The enhancement process can also alter the gradient distribution around keypoints, making SIFT descriptors less distinctive and more susceptible to incorrect matches. As a result, the technique may perform less effectively when applied to other datasets or in practical implementations.

Despite the advancements in biometric recognition, it is important to note that no specific geometry-based filtering methods have been identified for SIFT-based matching in existing literature for palm-vein recognition. However, in palmprint recognition, the SGR filter [12] has been proposed to enhance this matching process. The SGR filter performs extensive additional calculations by analysing ED and angles between matched keypoints within the same image and comparing them between the query and reference images. If the relative spatial relationships are consistent, the match is retained; otherwise, it is discarded.

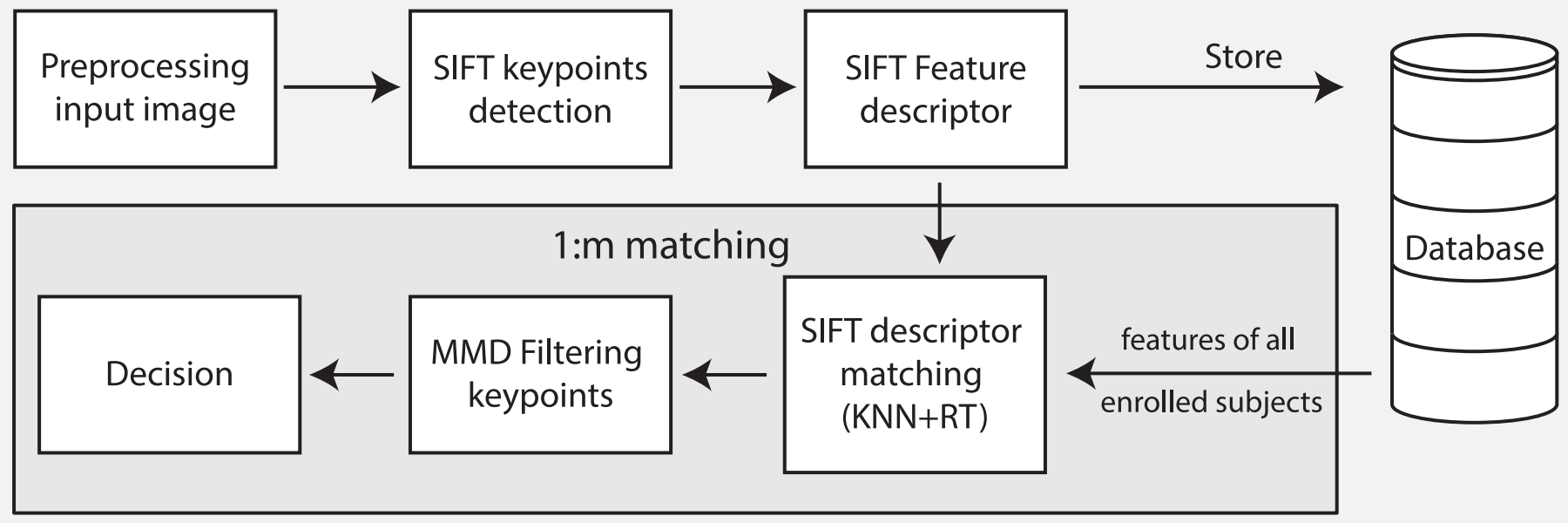

**Figure 4:** Flow graph of the proposed palm-vein recognition system using the MMD filter.

In contrast, the MMD filter only calculates the geometric distance between matched points using fewer computations and applies an algorithm to determine a match.

## 3. The Mean and Median Distance Filter (MMD)

The flow graph of the proposed palm-vein recognition system incorporating the MMD filter is presented in **Figure 4**.

The ROIs of the palm-vein images are first enhanced using the ILACS-LGOT contrast enhancement method. However, any contrast enhancement technique can be applied based on the specific requirements of the application or user preference. The extracted SIFT keypoints and descriptors are then stored in the database. At the feature matching stage, the pre-processing and feature extraction steps are repeated. The matched templates are then compared against the stored templates from the database and applied with the MMD filter.

The MMD filter is inspired by Lowe's [8] suggestion to use SIFT keypoints clusters for geometric fitting in object recognition through their affine projection. Various studies have adopted keypoints clustering methods with SIFT; for instance, Wang et al. [13] enhanced object recognition by using clusters of SIFT points, applying a seed points vector normalisation method derived from the vector normalisation used in the SIFT descriptor.

Similarly, Wang et al. [14] implemented SIFT point clustering in mammogram analysis for micro-calcification detection without preprocessing, achieving high sensitivity and specificity. Luo et al. [15] proposed using person-specific SIFT features and keypoints clusters in face recognition, demonstrating robustness against variations in expression, accessories, and pose. Fritz et al. [16] utilised clusters of SIFT points in processing UAV-acquired images to evaluate tree stem detection in open stands. However, as subtle changes in palm pose, finger or thumb positioning can introduce significant changes in the local area of a keypoint, using such point clusters are not suitable for palm-vein recognition.

The MMD filter is based on the principle that false positive match pairs have greater spatial distances compared to true positive matches, and that the $\mathcal{X}$ and $\mathcal{Y}$ coordinate differences of true matches should be minimal.

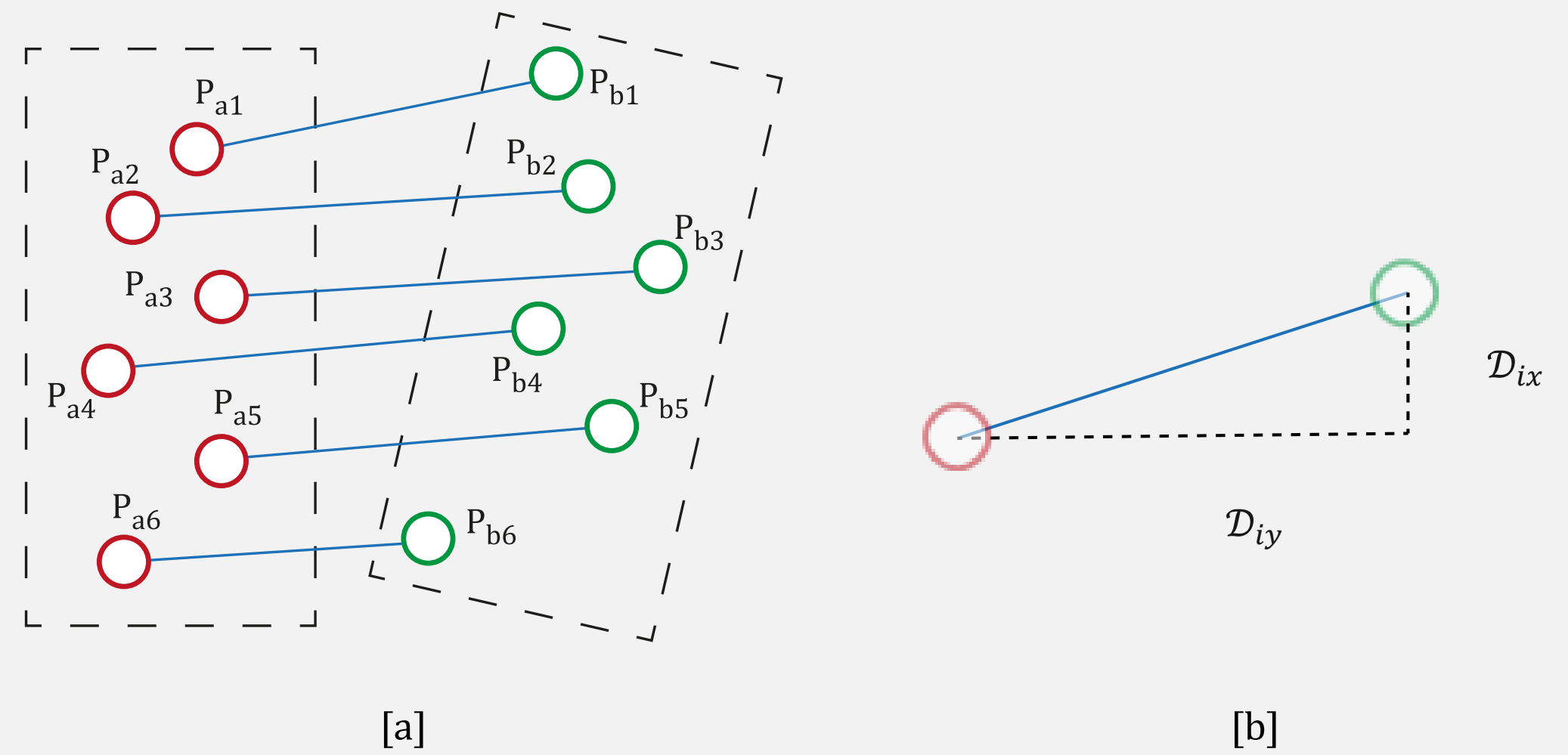

**Figure 5:**

[a] - Matched SIFT keypoints.

[b] - Horizontal ($\mathcal{D}_{ix}$) and vertical ($\mathcal{D}_{iy}$) distances between each pair of keypoints are calculated

When two palm-vein images are perfectly pre-aligned, true positive matches should ideally share the same $\mathcal{X}$ and $\mathcal{Y}$ coordinates, while false positives appear at different coordinates with a nonzero distance. However, variations in hand positioning, image noise, rotation, and scale introduce discrepancies, meaning even true positive matches will have slight coordinate differences.

Unlike the ED used in SIFT feature matching (which measures similarity in 128-dimensional vectors), the MMD filter focuses on the geometric placement of SIFT keypoints. It measures the spatial distance in pixels between $\mathcal{X}$ and $\mathcal{Y}$ coordinates of matching keypoints to filter out false matches **(Figure 5** [b]).

## 3.1. The MMD Algorithm

When matching SIFT features between two palm-vein images $\mathcal{G}$ and $\mathcal{P}$, for every matched keypoint pair ($\mathcal{M}_i$), measure the horizontal ($x$) and vertical ($y$) distances ($\mathcal{D}_{ix}$, $\mathcal{D}_{iy}$) between their coordinates. Then, calculate the respective mean ($\mu_x$, $\mu_y$) and median ($\tilde{x}_x$, $\tilde{x}_y$) distances on both axes.

Then, count the respective number of match pairs when both of these distances are below or equal to the mean ($\mathcal{D}_{ix} \leq \mu_x$ **and** $\mathcal{D}_{iy} \leq \mu_Y$) as $\mathcal{N}_L$, and when both of these distances are above their respective mean ($\mathcal{D}_{ix} > \mu_x$ **and** $\mathcal{D}_{iy} > \mu_y$) as $\mathcal{N}_H$. Two thresholds are introduced to accommodate rotation and scale variations of the images as a maximum mean threshold ($T_\mu$) and a maximum distance threshold ($T_\mathcal{D}$).

To consider if a pair of palm-vein images is likely to belong to the same subject, one of the following conditions should be met. If not, the algorithm assumes that there are 0 positive matches and rejects the image as a negative match.

- The total number of match pairs when their horizontal and vertical distances are below or equal to the mean, are higher than or equal to that of the total number of match pairs when these distances are above their respective mean ($\mathcal{N}_L \geq \mathcal{N}_H$).

- The horizontal and vertical mean values are equal to or below their respective thresholds ($\mu_x \leq T_\mu$ **and** $\mu_y \leq T_\mu$).

- The horizontal and vertical median values are equal to or below their respective mean distances ($\tilde{x}_x \leq \mu_x$ **and** $\tilde{x}_y \leq \mu_y$).

If the images meets the above selection criterion, all the following conditions should be met to determine if a pair of keypoints is a true positive match.

- Horizontal distance is lower than the horizontal mean and the threshold ($\mathcal{D}_{ix} < \mu_x$ **and** $\mathcal{D}_{ix} < T_{\mathcal{D}}$).

- Vertical distance is lower than the vertical mean and the threshold ($\mathcal{D}_{iy} < \mu_Y$ **and** $\mathcal{D}_{iy} < T_{\mathcal{D}}$).

**Algorithm 1** presents the pseudo-code for the MMD filter.

Algorithm 1: MMD Filter

**Input:** $\mathcal{X}, \mathcal{Y}$ coordinates of the matching feature points

Output:

1: for each $\mathcal{M}_i$

  1.1: $\mathcal{D}_{ix} \leftarrow \mathcal{X}_{\mathcal{G}} - \mathcal{X}_{\mathcal{P}}$

  1.2: $\mathcal{D}_{iy} \leftarrow \mathcal{Y}_{\mathcal{G}} - \mathcal{Y}_{\mathcal{P}}$

2: end for

3: $\mu_x \leftarrow mean\ of\ \mathcal{D}_{ix}$

4: $\mu_y \leftarrow mean\ of\ \mathcal{D}_{iy}$

5: $\tilde{x}_x \leftarrow median\ of\ \mathcal{D}_{ix}$

6: $\tilde{x}_y \leftarrow median\ of\ \mathcal{D}_{iy}$

7: $\mathcal{N}_L \leftarrow total\ number\ of\ matches\ when\ \mathcal{D}_{ix} \leq \mu_x\ and\ \mathcal{D}_{iy} \leq \mu_Y$

8: $\mathcal{N}_H \leftarrow total\ number\ of\ matches\ when\ \mathcal{D}_{ix} > \mu_x\ and\ \mathcal{D}_{iy} > \mu_y$

9: **if** $(\mathcal{N}_L \geq \mathcal{N}_H)$ **or** $(\mu_x \leq T_\mu$ **and** $\mu_y \leq T_\mu)$
  **or** $(\tilde{x}_x \leq \mu_x$ **and** $\tilde{x}_y \leq \mu_y)$ **then**

  9.1: for each $\mathcal{M}_i$ do

    9.1.1: **if** $(\mathcal{D}_{ix} < \mu_x)$ **and** $(\mathcal{D}_{ix} < T_{\mathcal{D}})$
      **and** $(\mathcal{D}_{iy} < \mu_Y)$
      **and** $(\mathcal{D}_{iy} < T_{\mathcal{D}})$ **then**

      9.1.1.1: accept $\mathcal{M}_i$ as a true positive

    9.1.2: **else** reject match

    9.1.3: end if

  9.2: end for

10: **else** reject the entire image

11: end if

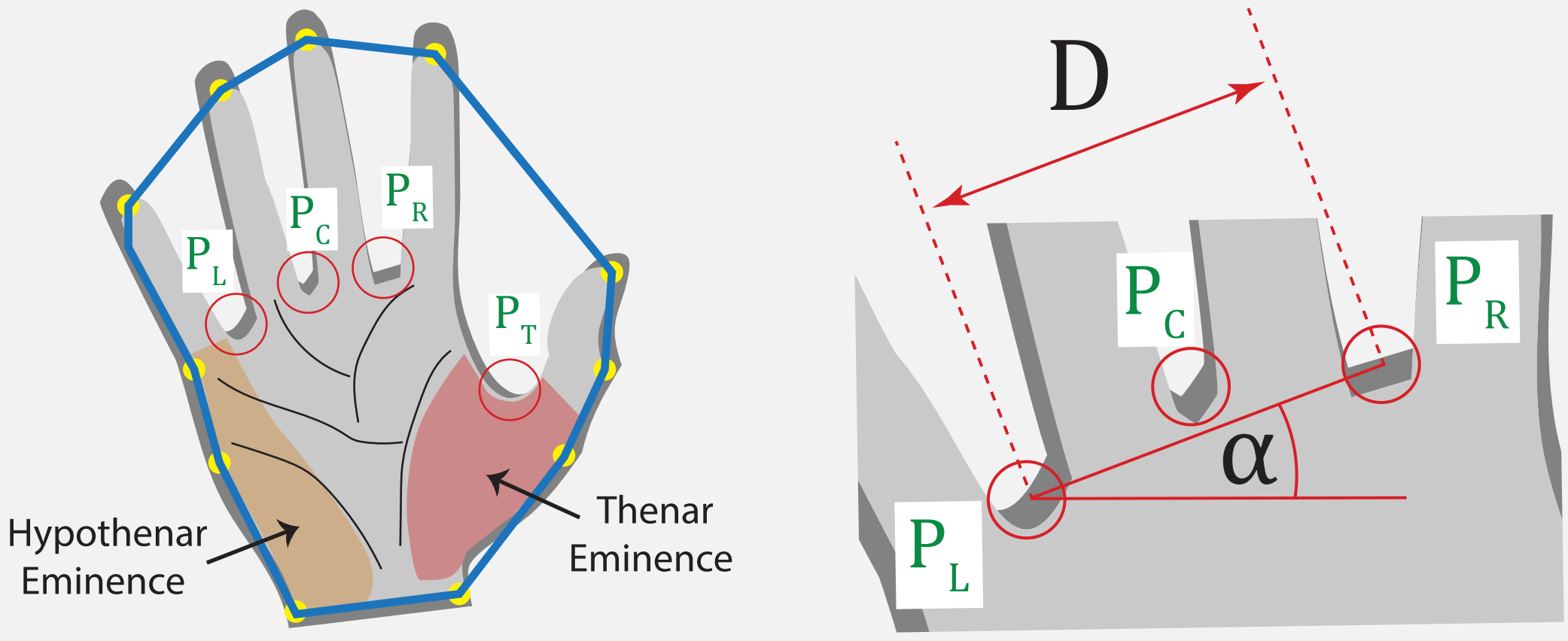

[a] [b]

**Figure 6:**
[a] - Palm valley point detection method as described in [136]. The image is first applied with OTSU thresholding and Suzuki contour detection. Then Sklansky's convex hull algorithm is used to identify the fingertips and the valley points.
$P_L$, $P_C$, $P_R$, and $P_T$ refers to left, centre, right, and thumb valley points, respectively.

[b] - Calculating the rotation angle 'α' and the distance 'D' between the left and right valley points.

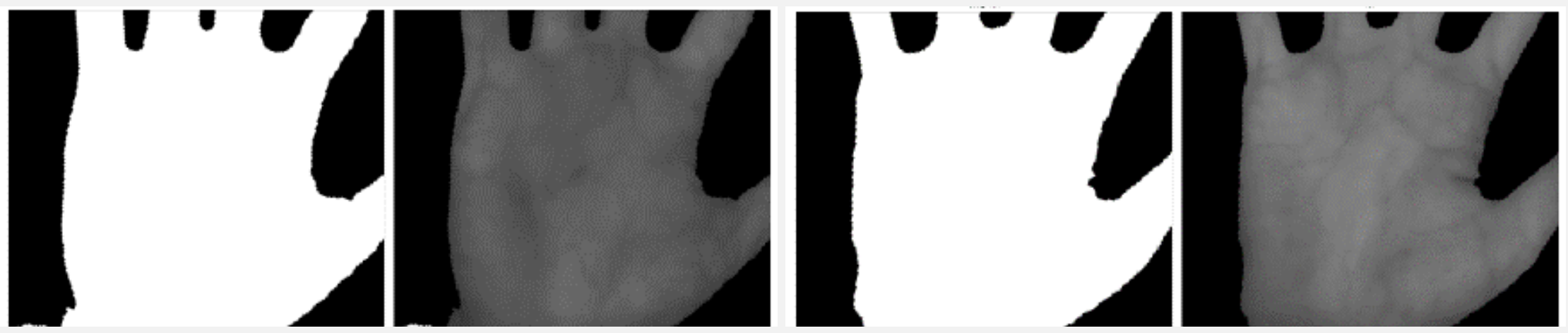

[a] [b]

**Figure 7:** Extracted ROI images and the corresponding mattes/masks.

## 4. Evaluation Methodology

The evaluation of the MMD filter is presented in two stages. This section details the experimental setup, including the dataset, the selection of the region of interest (ROI), and the feature matching and filtering techniques used for performance comparison. The experimental settings remain consistent with our previous work, ILACS-LGOT method in [4], [5], to ensure a fair comparison.

Previous studies on palm-vein identification often restricted their algorithm to a smaller region of interest (ROI) [17], [18], [19], [20], typically focusing on the central palmar area. However, visual examination of available datasets indicated that most palm veins are concentrated around the thenar eminence (from the base of the thumb to the wrist), the hypothenar eminence (from the base of the little finger to the wrist), and the upper palmar regions (see **Figure 6** [a]).

To maximise the available image data, in [4], [5] the entire palm was utilised instead of a limited ROI, following similar approaches to previous studies [21], [22], [23]. The valley points of the palm images were identified using the method described in [38], which employs Sklansky's convex hull algorithm, OTSU thresholding, and Suzuki contour detection, also producing a binary matte of the palm (see **Figure 6** [a]).

Next, the angle between the left and right valley points, located between the fingers, was calculated, and the image was rotated to align these valley points with the horizontal axis (see **Figure 6** [b]). The distance between these points, denoted as 'D' in **Figure 6** [b], was used as a unit of measurement to extract a 2D×2D ROI image centred on the palm, resulting in ROI images of varying sizes.

An eroded copy of the binary matte was then applied as a mask to filter out feature points near the edges (**Figure 7**). To enhance processing efficiency and reduce computational time, all images were resized to 60% of their original dimensions before further processing. Then the contrast enhancement method, ILACS-LGOT, was applied using 16×16 pixels image tiles to maintain consistency.

In the first stage, an evaluation of SIFT keypoint matching and parameter analysis is conducted. This stage demonstrates the variations between palm-vein image pairs and how the MMD filter parameters can be adjusted to accommodate them. Visual samples with various parameter combinations are presented.

The second stage involves a performance comparison using the same experimental settings as in our previous work [4], [5], for the ILACS-LGOT method, using EER as the performance measure. In that, the results were compared with existing SIFT and RootSIFT-based palm-vein recognition systems by substituting their image enhancement methods with the proposed ILACS-LGOT method. The stage two MMD filter evaluation compares the results by incorporating the MMD filtering step or replacing the existing filtering method with the MMD filter in the systems evaluated in [4], [5].

In this experimental setup, the CASIA dataset was divided to that 240 images (20%) were designated for testing during the algorithm development phase, while 960 images (80%) were used for evaluation in the 1:*m* closed-set approach. Consistent with the procedures outlined in previous experiments, left and right hands were treated as separate entities to maximise the sample size, though palms from the same subject were not cross-matched. In stage two, the MMD filter thresholds were set at 25 and 30 pixels for $T_\mu$ and $T_D$, respectively, to enhance feature point matching accuracy and minimise false positives.

# 5. Results and Analysis

## 5.1 Variations in Palm-Vein Images and MMD Parameters

**Figure 3** demonstrated SIFT+KNN+RT using various distance ratios. It was observed that when using higher RT values, a significant number of false matches were filtered in, while a lot of positive matches were filtered out with smaller RT values.

**Figure 8** demonstrates the results of SIFT+KNN+RT using two high RT values, with and without applying the MMD filter. A common threshold of 20 pixels was used for $T_\mathcal{D}$ and $T_\mu$. It can be observed from subfigures [b] and [d] of **Figure 8** how the MMD filter can effectively filter out false matches.

The thresholds $T_\mu$ and $T_D$ can be adjusted to control the sensitivity of the MMD filter and to accommodate scale and rotational variances between the images. **Figure 9** demonstrates the results of applying various thresholds from 10 to 35 pixels with increments of 5 pixels at each step. A common threshold was used for $T_\mathcal{D}$ and $T_\mu$ ($T_\mathcal{D} = T_\mu$).

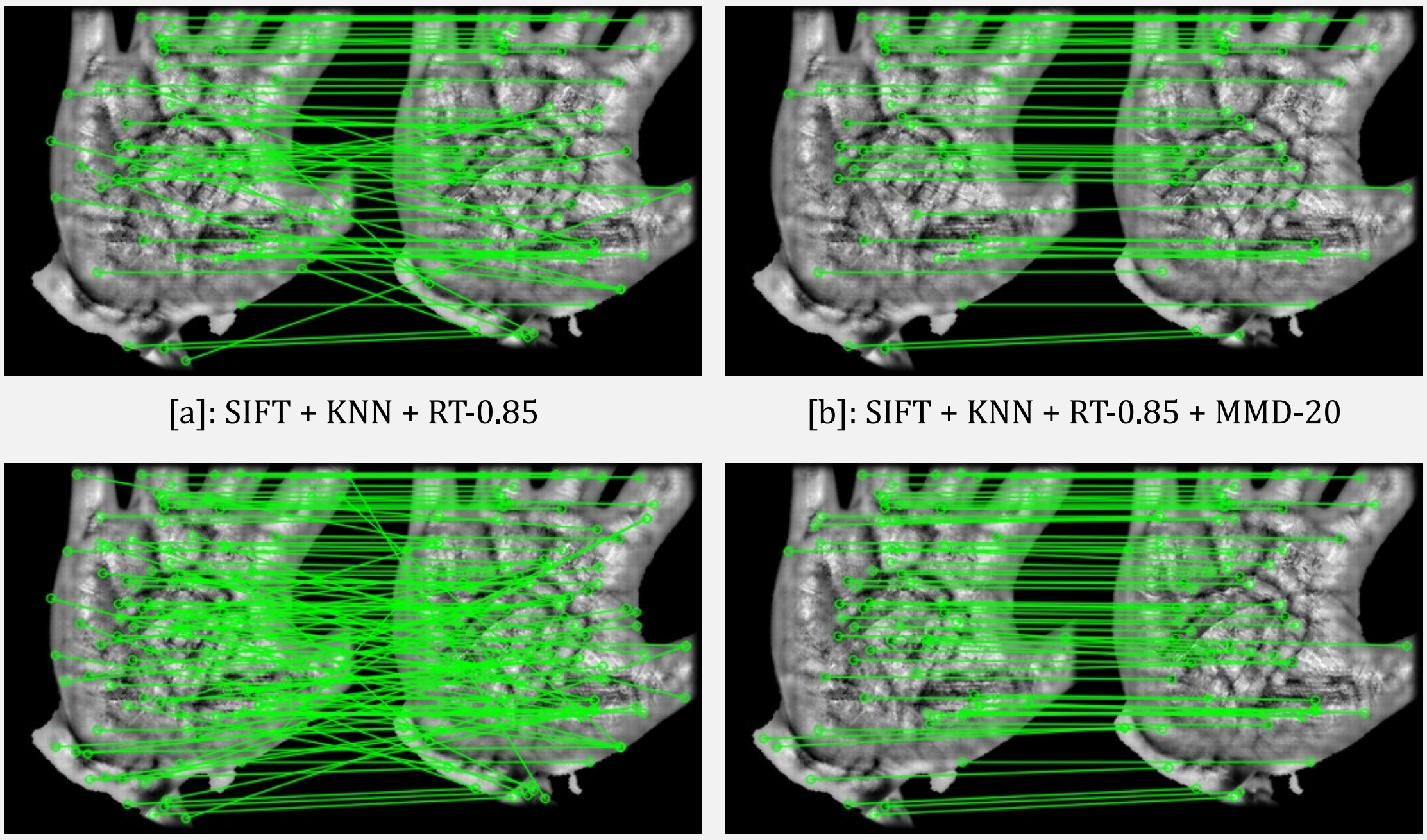

[a]: SIFT + KNN + RT-0.85
[b]: SIFT + KNN + RT-0.85 + MMD-20
[c]: SIFT + KNN + RT-0.9
[d]: SIFT + KNN + RT-0.9 + MMD-20

**Figure 9:** Comparison between before and after applying the MMD filter using a common threshold of 20 pixels $\left(T_{\mathcal{D}} = T_{\mu}\right)$.

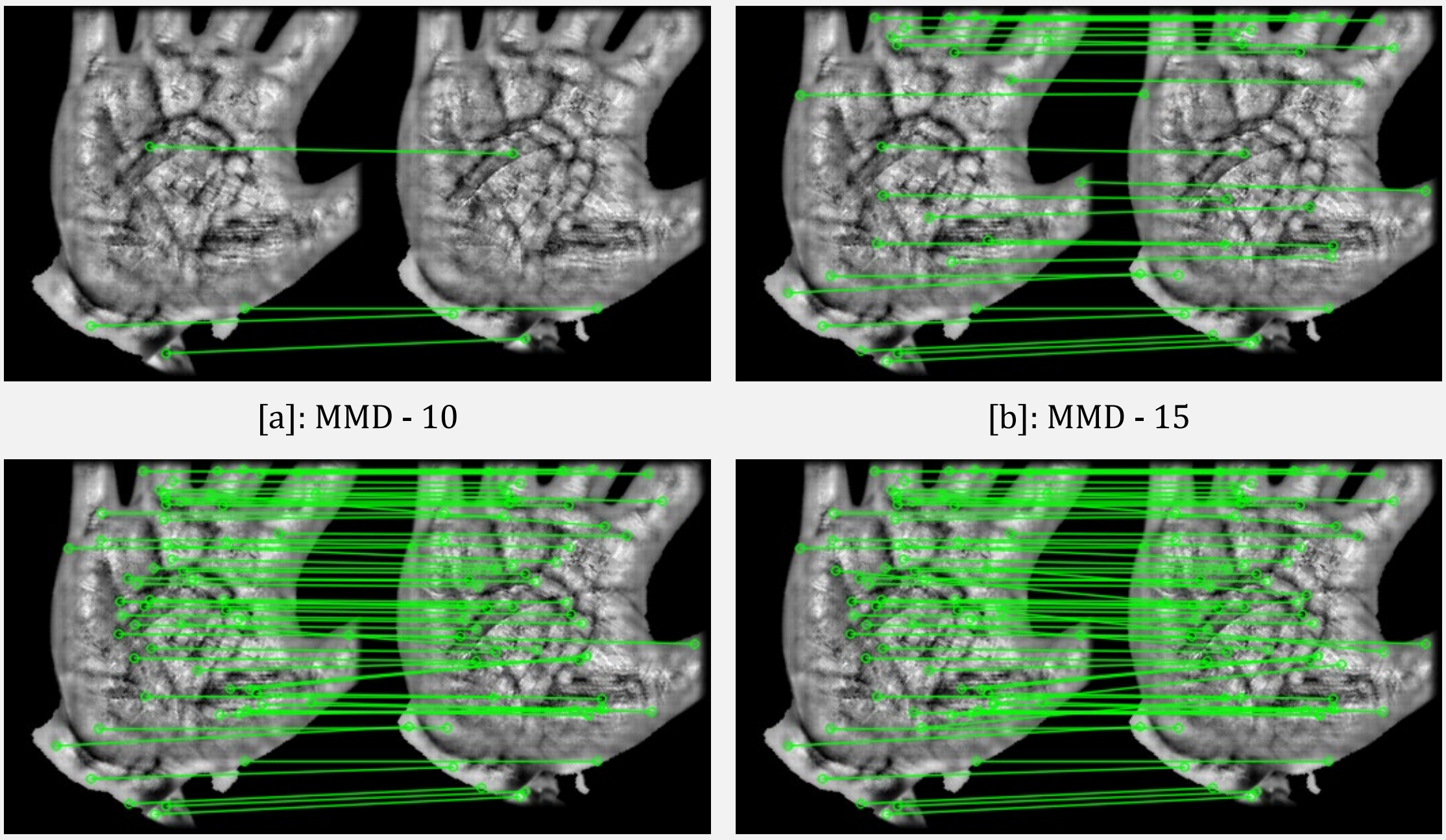

[a]: MMD - 10
[b]: MMD - 15
[c]: MMD - 25
[d]: MMD - 35

**Figure 8:** The effect of the MMD threshold ($T_{\mathcal{D}} = T_{\mu}$) with KNN + RT-0.9 + MMD filtering. Higher MMD thresholds may allow false SIFT matches to pass through the filter, while lower thresholds might filter out genuine positive SIFT matches.

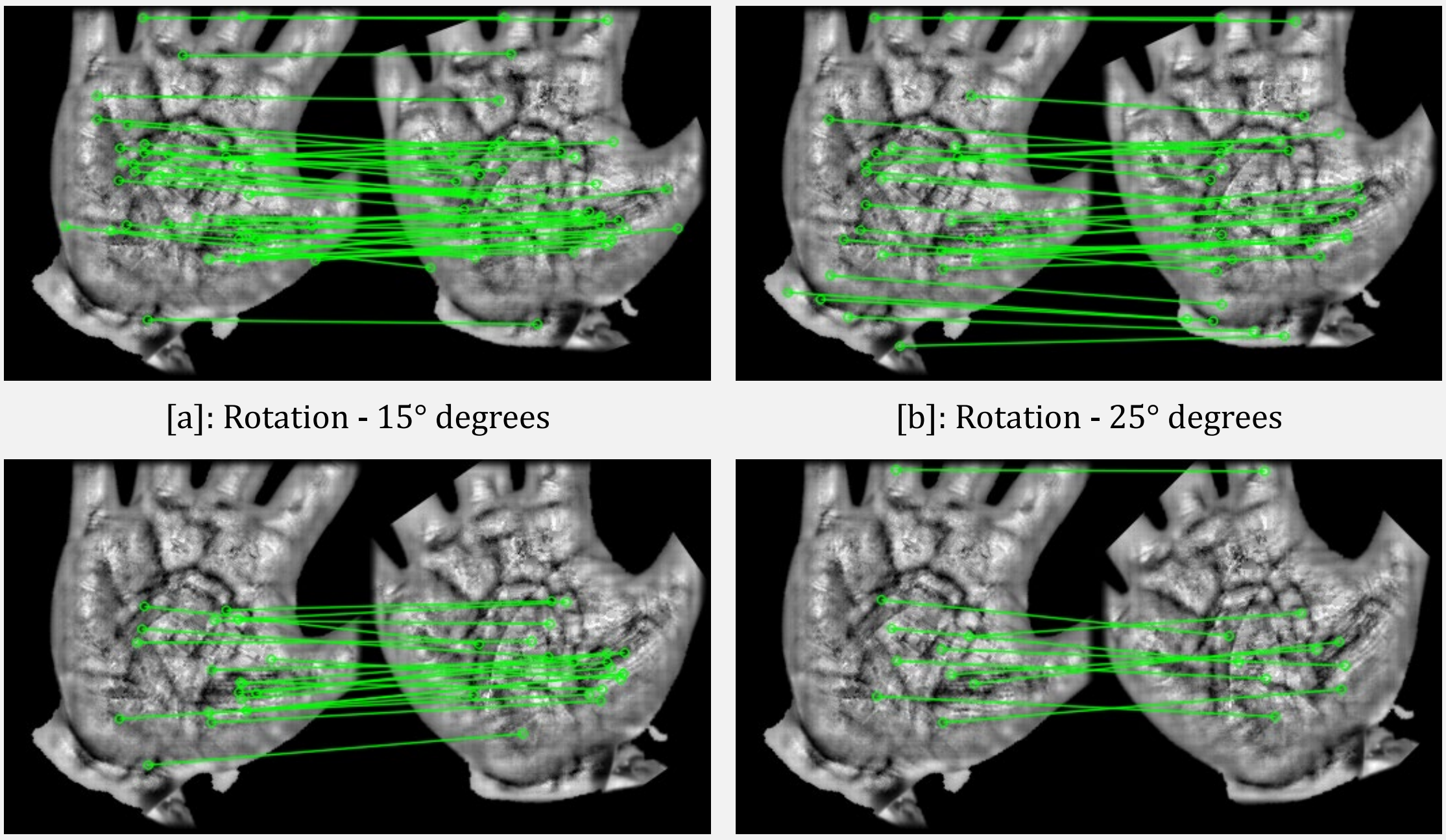

[a]: Rotation - 15° degrees

[b]: Rotation - 25° degrees

[c]: Rotation - 35° degrees

[d]: Rotation - 45° degrees

**Figure 11:** MMD filter can account for the rotational differences between the image pairs. KNN + RT-0.9 + MMD-30 ($T_{\mathcal{D}} = T_{\mu}$).

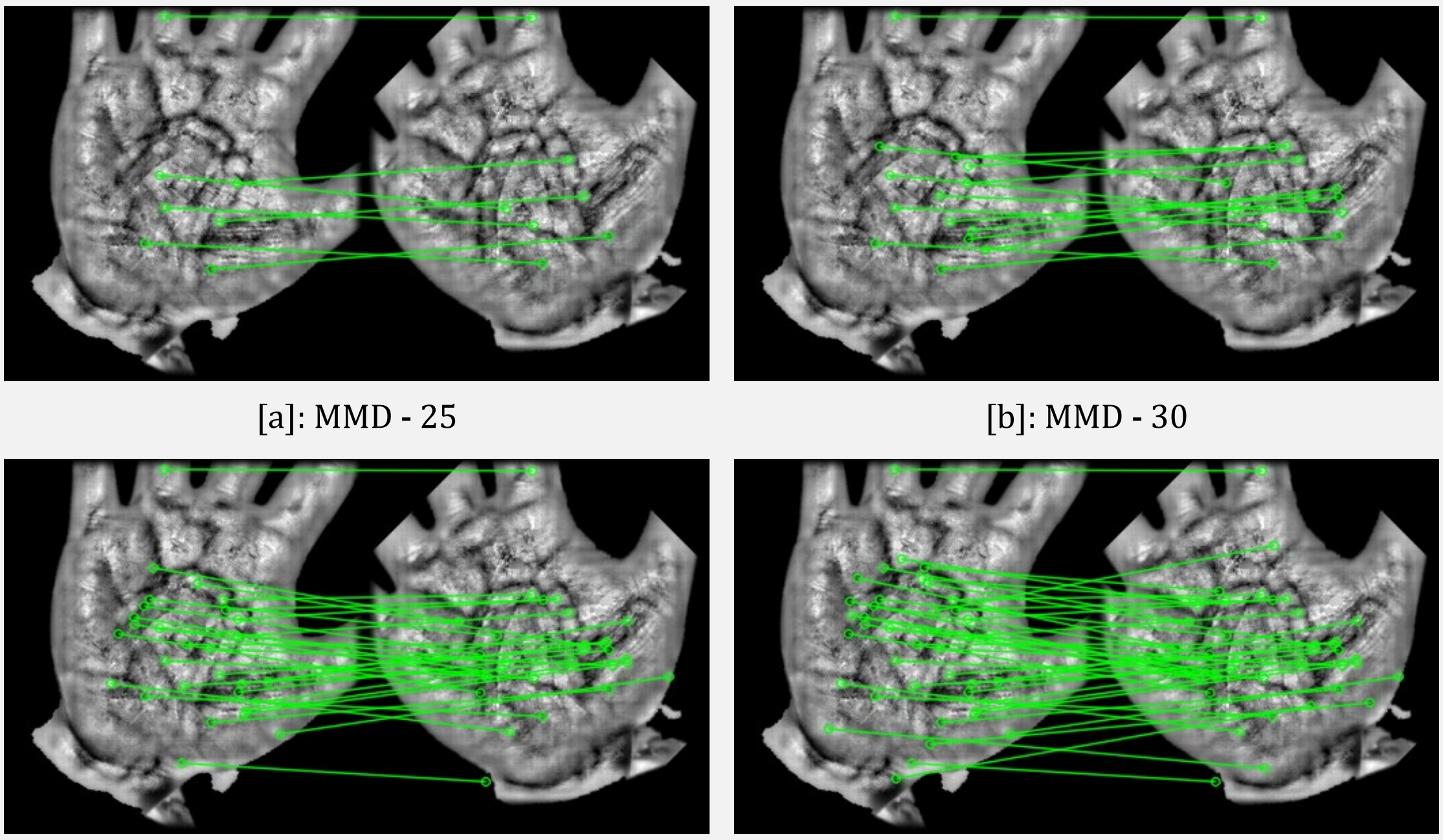

[a]: MMD - 25

[b]: MMD - 30

[c]: MMD - 35

[d]: MMD - 45

**Figure 10:** The thresholds can be adjusted to control the sensitivity of the MMD filter. KNN + RT-0.9 + MMD ($T_{\mathcal{D}} = T_{\mu}$), rotation - 45° degrees.

The MMD filter is designed to accommodate rotation variations of the palm-vein images.

**Figure 10** demonstrates the effect of introducing rotation to the right-hand side image of the match pair used for the demonstrations using a common threshold of 30 pixels.

**Figure 11** demonstrates how the threshold can be adjusted to reduce the sensitivity of the MMD filter to accommodate higher rotation variances.

## 5.1 Comparison and Verifying the Performance of the MMD filter

To assess the efficiency of the newly developed MMD filter, this section benchmarks its performance against established SIFT descriptor matching techniques using EER. The experimental setup was outlined in Section 4.

To find feature point pairs with minimum distances, Yan [24] applied a bidirectional feature matching methodology, where distance is measured between each feature point pair in both forward and backward directions. Acceptable matches are identified when ED is below predetermined thresholds, showing EERs of 0.65% for ORB and 1.84% for SIFT features, using 850nm CASIA dataset images.

In [23], KNN+RT was used with RootSIFT features and employed the bidirectional method from [24]. Results were reported with SIFT+ED+Bidirectional-matching. They used the 850nm images from the CASIA dataset with three sample images to produce the matching template. They further reported an EER value using ED. SIFT+KNN+RT and RootSIFT+KNN+RT+Bidirectional-matching were used to verify the performance of the ILACS-LGOT method in [4], [5]. However, our previous implementation of their bidirectional algorithm performed poorly. The experiments with ILACS-LGOT+SIFT+ED recorded an EER value of 3.33% and performed better than ILACS-LGOT+SIFT+ED+Bidirectional matching, which recorded an EER of 7.77%.

**Table 1:** Performance comparison with existing palm-vein recognition systems. Unless otherwise noted, the results are presented using a template size of 1.

| Recognition and filtering technique | EER % |
|---|---|
| (DoG-HE + SIFT) ED [20] (Template size: 3) | (Left hand) 2.87 |
| (ILACS-LGOT + SIFT) ED (Template size: 3) | (Left hand) 1.48 |
| (ILACS-LGOT + SIFT) ED + MMD (Template size: 3) | (Left hand) 1.12 |
| (CLAHE + block stretch + SIFT) ED + RANSAC [21] | 14.7 |
| (ILACS-LGOT + SIFT) ED + RANSAC [2] | 4.29 |
| (ILACS-LGOT + SIFT) ED + MMD | 3.01 |
| ECS-LBP + SIFT (ED) [22] | (L/R hands) 3.12/3.25 |
| (ILACS-LGOT + SIFT) ED [2] | (L/R hands) 2.75/2.88 |
| (ILACS-LGOT + SIFT) ED + MMD | (L/R hands) 2.46/2.62 |

**Table 2:** EER % values with SIFT, using template sizes of 1–5, compared with and without using the MMD filter.

| Template Size | ED [2] | ED+MMD | KNN+RT [2] | KNN+RT+MMD |
|---|---|---|---|---|
| 1 | 3.33 | 3.01 | 2.87 | 2.33 |
| 2 | 1.68 | 1.46 | 1.72 | 1.45 |
| 3 | 1.72 | 1.31 | 1.66 | 1.26 |
| 4 | 0.53 | 0.23 | 0.61 | 0.26 |
| 5 | 0.31 | 0.19 | 0.28 | 0.14 |

In [25], mismatches of SIFT+ED matching were removed with RANSAC. The reported EER value is 14.7%, and the AUC is 90.8% using the CASIA dataset. [26] presented SIFT+ED matching and used ECS-LBP as the image enhancement method. They separated the left and right-hand images into two subsets, reducing the number of interclass matches by a factor of 0.5, resulting in lower EER values [4]. The parameters used in the aforementioned systems were not discussed except for [23], where they used a threshold of 0.8 with RT when using RootSIFT features. The performance comparison of the palm-vein recognition systems is measured using EER and presented in **Table 1**.

The proposed MMD filtering method reduces EER values compared to the filtering methods in [25], [26], and [23]. The RANSAC filtering-based recognition method from [25] reported the highest EER value of 14.7%. Lowe [8] suggests that RANSAC filtering is not suitable when many outliers are present. The MMD filter was further tested with ED+MMD and KNN+RT+MMD using references with template sizes of 1–5. Results are presented in **Table 2** against ED and KNN+RT. The threshold used for RT in all the experiments is 0.7 [8]. It can be observed that EER values reduces as the template size is increased. The highest performance gain was observed between ED and ED+MMD.

When using a template size of one, ED+MMD reported a higher EER of 3.01% compared to 2.88% with KNN+RT. With template sizes of 2–5, ED+MMD reported better results than KNN+RT. The lowest EER values are recorded when using the KNN+RT+MMD method (0.14%), followed by the ED+MMD method (0.2%) with a template size of five.

In [4], [5], the ILACS-LGOT method outperformed the image enhancement methods compared to previous work. In this study, incorporating the MMD filter outperforms other filtering methods used in [4], [5] with the ILACS-LGOT method. This confirms that the MMD filter outperforms the existing filtering methods used for SIFT-based palm-vein recognition.

## 6. Conclusions

Palm vein recognition presents a significant challenge, as palm vein images can exhibit an infinite number of variations due to changes in hand posture caused by finger and thumb movements, as well as wrist rotation. The filtering methods currently employed in SIFT-based recognition systems continue to yield relatively high EER values.

This study introduces a novel filtering method, referred to as the Mean and Median Distance

(MMD) filter, designed to eliminate outliers or false positive matches in SIFT-based palm vein recognition. The MMD filter is highly robust to variations in scale and rotation. It operates on the principle that when two images are properly aligned and superimposed, the geometric distance between corresponding keypoints should be minimal. To accommodate scale and rotation variations, the MMD filter assesses the mean and median distances separately in both the horizontal and vertical directions. The sensitivity of the filter can be adjusted using two predefined thresholds.

Experiments were conducted using the MMD filter alongside the ED, KNN, and RT algorithms, with 1 to 5 registration template images. The results demonstrate that the proposed MMD filter outperformed other filtering methods applied to SIFT-based palm vein recognition. Furthermore, the MMD filter is computationally efficient, requiring minimal calculations. Notably, the MMD filter is not restricted to SIFT-based systems and can be applied to any feature-matching process to remove outliers. It considers only the median and mean distances of matching pairs, a method that could be further enhanced using machine learning techniques.